\documentclass[letterpaper, 10 pt, conference]{ieeeconf}  
\IEEEoverridecommandlockouts                              
\usepackage{flushend}
\usepackage{endnotes}
\usepackage{import}
\usepackage{amsopn}
\usepackage[table, x11names]{xcolor}
\usepackage{tikz,filecontents}
\usepackage{tabularx,booktabs} 
\usepackage[export]{adjustbox}
\usepackage[colorlinks=true,linkcolor=black,citecolor=black,urlcolor=black]{hyperref}
\usepackage{times} 
\usepackage{amsmath} 
\usepackage{amssymb}  
\usepackage{graphicx}
\usepackage{algorithm}
\usepackage{mathtools}
\usepackage{algorithm}
\usepackage{algorithmicx}
\usepackage{inputenx}
\usepackage[noend]{algpseudocode}
\usepackage[font=footnotesize]{caption}
\usepackage[font=footnotesize]{subcaption}
\let\labelindent\relax
\usepackage{enumitem}
\usepackage{textcomp}
\usepackage{epstopdf}
\usepackage{url}
\usepackage{float}
\usepackage{pgfplots}

\definecolor{brewerGreen0}{HTML}{E5F5F9}
\definecolor{brewerGreen1}{HTML}{99D8C9}
\definecolor{brewerGreen2}{HTML}{2CA25F}

\definecolor{brewerCyan0}{HTML}{ECE2F0}
\definecolor{brewerCyan1}{HTML}{A6BDDB}
\definecolor{brewerCyan2}{HTML}{1C9099}

\definecolor{brewerGrey0}{HTML}{F0F0F0}
\definecolor{brewerGrey1}{HTML}{BDBDBD}
\definecolor{brewerGrey2}{HTML}{636363}

\definecolor{revisionColor}{HTML}{0238A8} 
\definecolor{lastRevisionColor}{HTML}{CC4C02} 

\usepackage{lipsum}

\usepackage{cite}

\usepackage{balance}
\usepackage{multirow} 

\usepackage{booktabs}

\usepackage{glossaries-extra}
\setabbreviationstyle[acronym]{long-short}
\glssetcategoryattribute{acronym}{nohyperfirst}{true}

\makeglossaries

\newacronym{slam}{SLAM}{Simultaneous Localization and Mapping}
\newacronym{ba}{BA}{Bundle Adjustment}
\newacronym{sfm}{SfM}{Structure from Motion}
\newacronym{pgo}{PGO}{Pose-Graph Optimization}
\newacronym{vpr}{VPR}{Visual Place Recognition}
\newacronym{sgd}{SGD}{Stochastic Gradient Descent}
\newacronym{ils}{ILS}{Iterative Least-Squares}
\newacronym{gn}{GN}{Gauss-Newton}
\newacronym{lm}{LM}{Levenberg-Marquardt}
\newacronym{pcg}{PCG}{Preconditioned Conjugate Gradient}
\newacronym{map}{MAP}{Maximum-A-Posteriori}
\newacronym{gf}{GF}{Gaussian Filters}
\newacronym{pf}{PF}{Particle Filters}
\newacronym{sdp}{SDP}{Semi-Definite Programming}
\newacronym{vo}{VO}{Visual Odometry}
\newacronym{vio}{VIO}{Visual-Inertial Odometry}
\newacronym{lo}{LO}{LiDAR Odometry}
\newacronym{vlo}{VLO}{Visual-LiDAR Odometry}
\newacronym{imu}{IMU}{Inertial Measurement Unit}
\newacronym{ros}{ROS}{Robot Operating System}
\newacronym{ate}{ATE}{Absolute Trajectory Error}
\newacronym{ape}{APE}{Absolute Pose Error}
\newacronym{rpe}{RPE}{Relative Pose Error}
\newacronym{vins}{VINS}{Visual INertial System}
\newacronym{pod}{POD}{Plain Old Data}
\newacronym{dpc}{DPC}{Dynamic Property Container}
\newacronym{boss}{BOSS}{Basic Object Serialization System}
\newacronym{mpc}{MPC}{Model Predictive Control}
\newacronym{qr}{QR}{Quadratic Regulator}
\newacronym{al}{AL}{Augmented Lagrangian}
\newacronym{ip}{IP}{Interior Point}
\newacronym{sqp}{SQP}{Sequential Quadratic Programming}
\newacronym{teb}{TEB}{Timed Elastic Band}
\newacronym{admm}{ADMM}{Alternating Direction Method of Multipliers}
\newacronym{isqp}{iSQP}{Iterative Sequential Quadratic Programming}
\newacronym{nlp}{NLP}{NonLinear Programming}
\newacronym{ipopt}{IPOPT}{Interior Point OPTimizer}
\newacronym{sota}{SOTA}{State Of The Art}
\newacronym{svd}{SVD}{Singular Value Decomposition}

\def\slam{\gls{slam} }

\newacronym{kslam}{K-SLAM}{Keyframe-based \slam}

\def\figref#1{Fig.~\ref{#1}}
\def\tabref#1{Tab.~\ref{#1}}
\def\eqref#1{Eq.~(\ref{#1})}

\def\etal{\emph{et al.}}

\newcounter{todonum}

\hypersetup{draft}

\let\footnote\endnote
\usepackage{amsopn}

\newcommand{\bt}{\mathbf{t}}

\newcommand{\bA}{\mathbf{A}}

\newcommand{\bF}{\mathbf{F}}

\newcommand{\bG}{\mathbf{G}}
\newcommand{\bH}{\mathbf{H}}
\newcommand{\bI}{\mathbf{I}}
\newcommand{\bX}{\mathbf{X}}
\newcommand{\bZ}{\mathbf{Z}}
\newcommand{\bR}{\mathbf{R}}
\newcommand{\bS}{\mathbf{S}}
\newcommand{\bU}{\mathbf{U}}
\newcommand{\bV}{\mathbf{V}}

\newcommand{\bblambda}{\boldsymbol{\lambda}}
\newcommand{\bbmu}{\boldsymbol{\mu}}

\newcommand{\bJ}{\mathbf{J}}
\newcommand{\bZero}{\mathbf{0}}

\newcommand{\bb}{\mathbf{b}}

\newcommand{\be}{\mathbf{e}}

\newcommand{\ff}{\mathbf{f}}

\newcommand{\bg}{\mathbf{g}}
\newcommand{\bq}{\mathbf{q}}
\newcommand{\bP}{\mathbf{P}}

\newcommand{\bx}{\mathbf{x}}

\newcommand{\bz}{\mathbf{z}}
\newcommand{\bu}{\mathbf{u}}

\newcommand{\bDelta}{\mathbf{\Delta}}

\newcommand{\bDeltax}{\mathbf{\Delta x}}

\newcommand{\bzero}{\mathbf{0}}

\newcommand{\pluseq}{\mathrel{+}=}
\newcommand{\defeq}{=}

\newcommand{\bmu}{\boldsymbol{\mu}}

\newcommand{\bOmega}{\mathbf{\Omega}}

\DeclareMathOperator*{\argmin}{argmin}

\def\kf{{k_\mathrm{f}}}
\def\kg{{k_\mathrm{g}}}

\def\g2o{$g^2o$}
\def\se2{\mathrm{SE}(2)}
\def\se3{\mathrm{SE}(3)}
\def\sim3{\mathrm{S}(3)}
\def\so3{\mathrm{SO}(3)}
\def\me{\mathrm{e}}
\def\t2v{\mathrm{t2v}}
\def\v2t{\mathrm{v2t}}

\begin{document}
	\title{\LARGE \bf How-to Augmented Lagrangian on Factor Graphs}
	
	\author{Barbara Bazzana $^{1}$ \and Henrik Andreasson $^{2}$ \and Giorgio Grisetti $^{1}$
		\thanks{$^{1}$Barbara Bazzana and Giorgio Grisetti are with the Department of Computer,
			Control, and Management Engineering  ``Antonio Ruberti", Sapienza University of
			Rome, Rome, Italy {\tt\footnotesize{\{bazzana, grisetti\}@diag.uniroma1.it}}
		}%
	\thanks{$^{2}$Henrik Andreasson is with the Centre for Applied Autonomous Sensor Systems (AASS), \"Orebro University, \"Orebro, Sweden {\tt\footnotesize{henrik.andreasson@oru.se}}
}%
	}
\maketitle

\begin{abstract}
  %
Factor graphs are a very powerful graphical representation, used to model many problems in robotics. They are widely spread in the areas of \gls{slam}, computer vision, and localization. In this paper we describe an approach to fill the gap with other areas, such as optimal control, by presenting an extension of Factor Graph Solvers to constrained optimization. The core idea of our method is to encapsulate the \gls{al} method in factors of the graph that can be integrated straightforwardly in existing factor graph solvers. 

We show the generality of our approach by addressing three applications, arising from different areas: pose estimation, rotation synchronization and \gls{mpc} of a pseudo-omnidirectional platform. We implemented our approach using C++ and ROS. Besides the generality of the approach, application results show that we can favorably compare against domain specific approaches. 
\end{abstract}

\section{Introduction}
\label{sec:intro}

Nonlinear Optimization is at the core of many robotics applications across various fields, such as mobile robotics~\cite{andreasson2022mdpi}, \gls{slam}~\cite{ila2017ijrr, grisetti2012iros}, \gls{sfm}~\cite{schonberger2016cvpr} and calibration~\cite{cicco2016icra}.
The workflow consists of two stages. First, the variables to be computed are identified and the problem to be solved is modeled as a cost function. Such a function expresses the objectives to be achieved through relations involving the variables. Examples of such objectives can be: reaching the goal with  limited control inputs and avoiding obstacles, or finding the a-posteriori trajectory which is maximally consistent with the measurements received from the sensors. Once the problem is formalized, its solution is devolved to the most suitable optimizer. They differ based on the method they implement. Some of them are general-purpose, such as IFOPT~\cite{winkler2018ifopt}, others target at area-specific formulations, such as ACADOS~\cite{verschueren2021mpc} for optimal control. Factor graphs are widely used to both model and solve unconstrained nonlinear optimization problems, relying on \gls{ils} solvers, such as those developed in the field of \gls{slam}~\cite{grisetti2020solver}. In this paper, we present the \gls{al}-extension of~\cite{grisetti2020solver} to constrained optimization, leveraging on recent results from the work of Sodhi \etal~\cite{sodhi2020icra}, Qadri \etal~\cite{qadri2022incopt} and ours~\cite{bazzana2023ral}. 

\begin{figure}[t]
	\begin{center}
	\begin{tabular}{cc}
			\includegraphics[width=0.48\columnwidth,valign=c]{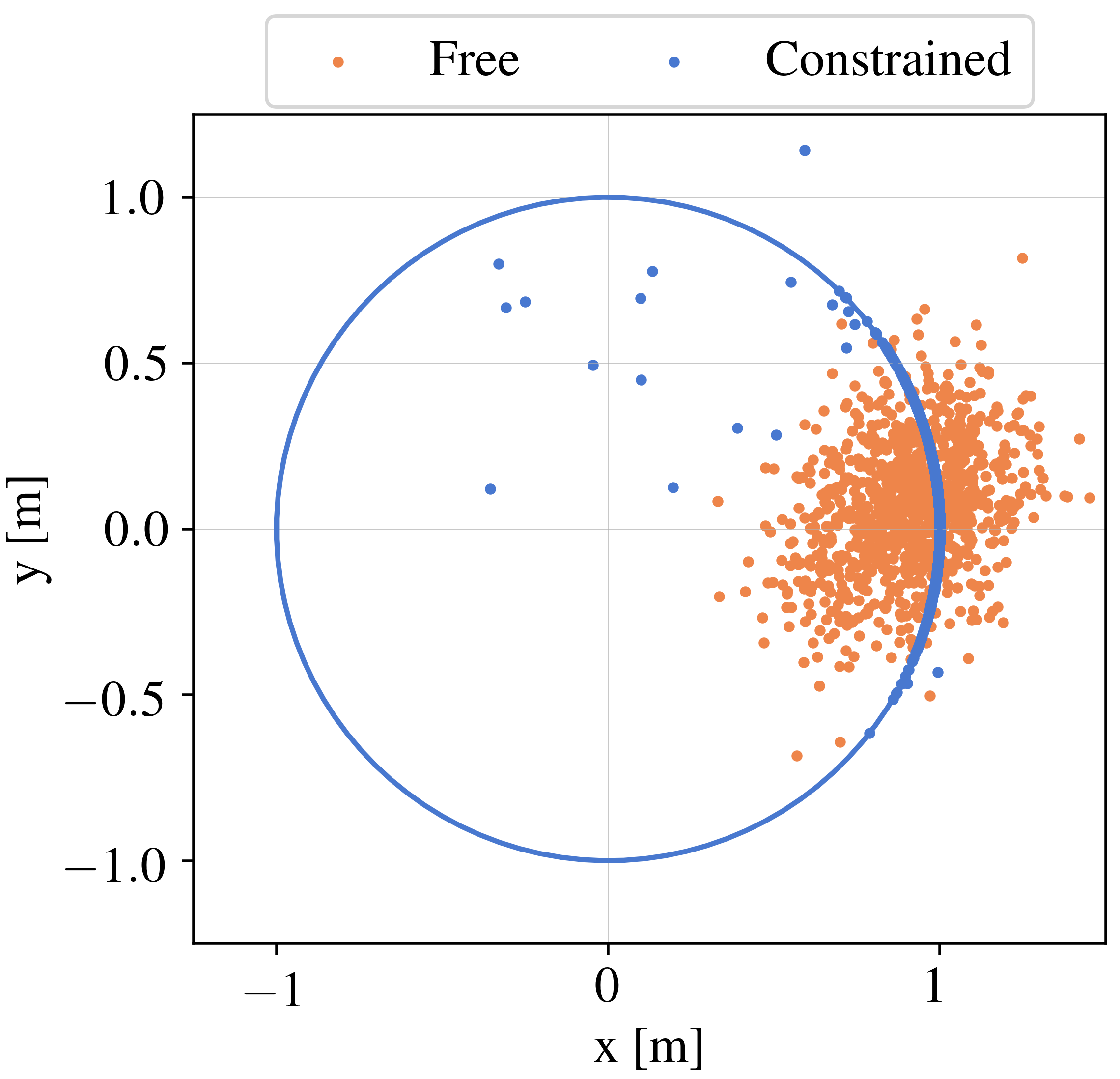}&
			\includegraphics[width=0.45\columnwidth,valign=c]{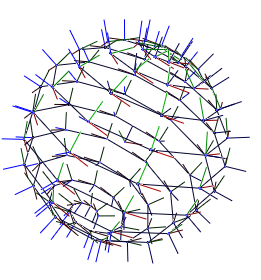}\\
			\centering(a)& \centering(b)
		\end{tabular}
	\end{center}
		\centering
		\includegraphics[width=0.65\columnwidth]{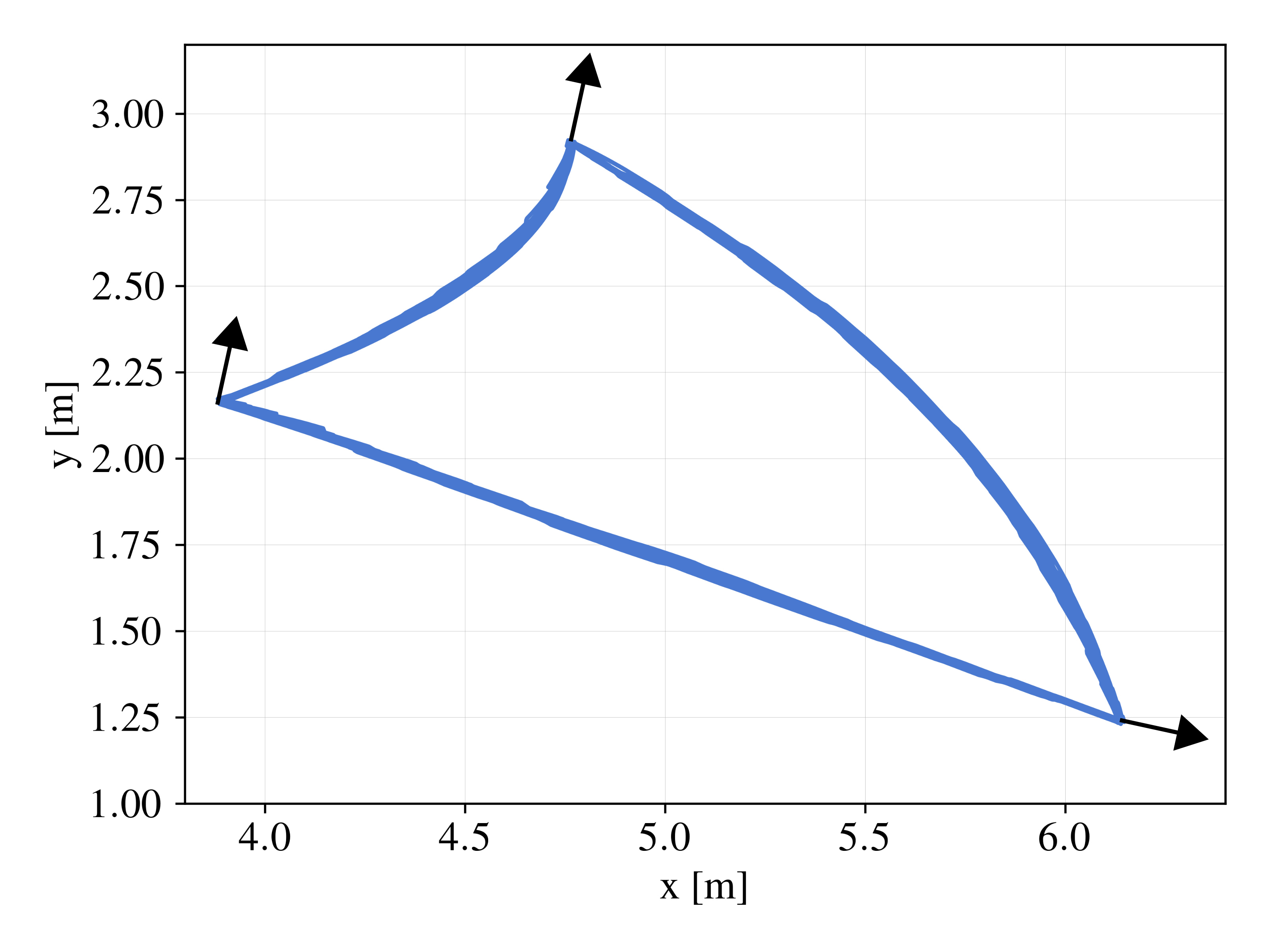} \\
		\centering
		(c)
		\centering
		\caption{Three application studies: (a) pose estimation; (b) rotation synchronization; (c) \gls{mpc} of a pseudo-onidirectional platform; the robot is traveling across three goals, with desired orientation represented by the arrows.}
		\vspace{-20px}
		\label{fig:motivation}
	\end{figure}

The core idea of our method is to use the \gls{al} method to model a new type of factors which can be directly included in existing unconstrained solvers: the constraint factors. Handling constraints enlarges the application domain of factor graphs confirming them as a general framework for optimization in robotics. 

In this paper, we first review the theoretical bases and subsequently present an algorithm scheme that might be used as reference implementation. 
We cast this algorithm to three increasingly complex problems, which are shown in \figref{fig:motivation}. The first is pose estimation of a unicycle, subject to the constraint that the estimate is coherent with the kinematics model. In this example  the robot starts in the origin and applies only linear velocity, therefore it is known to lye on the circumference independently from the initial unknown orientation. The second is rotation synchronization~\cite{eriksson2021tpa} subject to the constraint that the estimate is actually a rotation matrix. The system can recover orientations of the poses from arbitrary initial guesses. The third is \gls{mpc} of the pseudo-omnidirectional platform presented in~\cite{andreasson2022mdpi}, which comprises dynamics, velocity and acceleration constraints. In introducing each application, we provide the reader with practical insights on implementation and parameter choices. With these three examples, we show that using constrained factor graphs and \gls{al} can produce results that compare favorably against domain specific approaches.



\section{Related Work}
\label{sec:related}
Factor graphs optimization is a very powerful tool to compute optimal solutions to many problems in robotics~\cite{dellaert2021ar}. Factor graph solvers exploit the sparsity pattern for efficiency. They address nonlinear unconstrained optimization problems using \gls{ils}~\cite{grisetti2020solver}. Extending factor graphs to constrained optimization is a relevant topic: it allows both new ways of addressing old problems, such as distributed or robust \gls{slam}~\cite{cunningham2015icra, choudhary2015icra, bai2018iros}, and new applications of the tool, such as Optimal Control~\cite{xie2020corr, sodhi2020icra, qadri2022incopt}. 

Choudhary \etal~\cite{choudhary2015iros} address the problem of memory efficiency in \gls{slam} by splitting the graph into sub-graphs and imposing consistency of the separators using hard constraints. The resulting optimization problem is solved in a decentralized manner using the multi-block \gls{admm}~\cite{boyd2011ml}. Differently, Cunningham \etal~\cite{cunningham2010iros} use Gram-Schmidt orthogonalization for elimination of the constrained variables when solving linear constrained sub-problems. In order to boost robustness against local minima in \gls{slam}, Bai \etal~\cite{bai2018iros} represent loop closures as constraints and use \gls{isqp} to solve the resulting constrained \gls{slam} graphs.

The versatility of factor graphs was exploited to address motion planning problems~\cite{dong2016rss, mukadam2019ar} which increased the interest in constraints-embedding factor graphs. Yang \etal~\cite{yang2021icra} propose to devolve the solution of the constrained optimization problem for variable elimination to a specialized solver. They then focus on the Linear Quadratic Regulator problem,  where the constrained sub-problem can be trivially solved. The difference with our method is that our factor graph-solver embeds general constraints, without the need of relying on a specialized solver. Also \gls{isqp} was investigated as a method to embed nonlinear constraints in factor graph-based estimation and \gls{mpc} on Unmanned Aerial Vehicles by Ta~\etal~\cite{ta2014icuas}. Finally, Xie~\etal~\cite{xie2020corr} convert the constrained problem into an unconstrained one by introducing a loss function for each constraint. They present motion planning applications ranging from cart-pole to quadruped robots. 

Orthogonal to~\cite{xie2020corr, ta2014icuas}, the method proposed by Sodhi \etal~\cite{sodhi2020icra} leverages on the \gls{al} method to extend the incremental smoothing solver by Kaess \etal~\cite{kaess2008tro} with constraints-handling. They explain how to represent constrained optimization  over a Bayes Tree. More recently, the newer version of the solver was presented by Qadri \etal~\cite{qadri2022incopt} where online relinearization is used for efficiency. In the computer vision literature, the \gls{al} method was adopted by Eriksson \etal~\cite{eriksson2021tpa} to address the rotation synchronization problem. We present it here as well, with a focus on its embedding in our framework.

Inspired by~\cite{sodhi2020icra, qadri2022incopt}, this work revisits \gls{al} on factor graphs and provides an implementation scheme of our primal-dual procedure. We test its generality with application to three problems from different areas and of varying complexity. The \gls{mpc} application is supported with real-world experiments. Moreover, we comment on the adaptation schemes used for the main parameters. This work builds on our previous work~\cite{bazzana2023ral} and generalizes its ideas, including general nonlinear constraints in the formulation. Finally, we here propose a different \gls{al} function from the one previously used in the literature of factor graphs, and compare the two.

\section{Our Approach}
\label{sec:main}
In this paper we present an extension of \gls{ils} on factor graphs to solve \gls{nlp} using the \gls{al} method. Factor graphs are bipartite graphs with two kinds of nodes: variables and factors. Variables represent the state of our system, while factor nodes model dependence relationships between the neighbor variables.

Let $\bx=\bx_{0:N-1}$ be the set of all variables which can span over arbitrary continuous domains, with $\bx_n$ element of a manifold~\cite{hertzberg2013sd}, e.g. the special Euclidean group $\se3$. Let us represent the $k^\mathrm{th}$ factor as $\left<\bOmega_k, \be_k(\cdot) \right>$  with $\be_k\in\mathbb{R}^{z}$ representing the difference between predicted and actual $k^\mathrm{th}$ measurement with information matrix $\bOmega_k\in\mathbb{R}^{z\times z}$, only depending on the subset of variables $\bx^k=\bx_{k_0:k_K}$. 
A factor graph models the summation
\begin{eqnarray}
	\label{eq:fg_log_likelihood}
	F(\bx)=\sum_{k =0} ^{K-1}||\be_k(\bx^k)||_{\bOmega_k}^2
\end{eqnarray}
where $||\be_k(\bx^k)||_{\bOmega_k}^2 \defeq \be_k(\bx^k)^T \bOmega_k \be_k(\bx^k)$. Under Gaussian assumptions, ~\eqref{eq:fg_log_likelihood} expresses the negative log-likelihood of the measurements given the states. Factor graph-solvers compute the variables $\bx$ which minimize~\eqref{eq:fg_log_likelihood}, using the \gls{ils} approach~\cite{grisetti2020solver}. At each iteration, the current solution $\hat \bx$ is refined by taking a Gauss-Newton step over \eqref{eq:fg_log_likelihood}: $\hat \bx \leftarrow \hat\bx \boxplus \bDeltax$; where $\bH \bDeltax = -\bb$, and $\boxplus$ adds the Euclidean perturbation $\bDeltax$ to $\hat\bx$ in the manifold space. By using the first order Taylor expansion of the error function $\be_k(\hat{\bx}\boxplus\bDelta\bx) \simeq \hat{\be}_k+\bJ_k\bDelta\bx$ around $\hat\bx$ in \eqref{eq:fg_log_likelihood}, we get
\begin{equation}
	\label{eq:linear-sys-unconstrained}
	\bb=\sum_{k =0} ^{K-1}\overbrace{\bJ_k^T\bOmega_k\hat{\be}_k}^{\bb^k} \qquad
	\bH=\sum_{k =0} ^{K-1}\overbrace{\bJ_k^T\bOmega_k\bJ_k}^{\bH^k}
\end{equation} 

\subsection{Augmented Lagrangian for Nonlinear Programming}
Consider the following \gls{nlp} problem
\begin{eqnarray}
	\label{eq:nlp}
&\mathbf{x}^*=\mathop{\argmin}\limits_{\bx}\overbrace{\sum_{k=0}^{K-1}||\be_k(\bx^k)||_{\bOmega_k}^2}^{F(\bx)},\\\nonumber \mathrm{subject\,to}\,& \ff_{\kf}(\bx^{\kf})=\bZero,\,\kf=0,...,K_\mathrm{f}-1,\\\nonumber& \bg_{\kg}(\bx^{\kg})\leq\bZero,\,\kg=0,...,K_\mathrm{g}-1
\end{eqnarray}
with $K_\mathrm{f}$ multidimensional equality constraints $\ff_{k_f}(\cdot)$ and $K_\mathrm{g}$ multidimensional inequality constraints $\bg_{\kg}(\cdot)$, only involving a subset of variables, respectively $\bx^{\kf}$
and $\bx^{\kg}$.  \eqref{eq:nlp} can be converted into an equality constrained problem by introducing $K_{\mathrm{g}}$ vectors of the type $\bq_{\kg}=(q_0\,,q_1\,,\,...,\,q_{\kg,J-1})^T$ with $q_i>0,\,\forall\, i=0,1,...,\kg,J-1$, one for each inequality constraint $\bg_{\kg}(\cdot)$ of dimension $\kg,J$
\begin{eqnarray}
	\label{eq:nlp-equality}
	&\mathbf{x}^*=\mathop{\argmin}\limits_{\bx}\overbrace{\sum_{k=0}^{K-1}||\be_k(\bx_k)||_{\bOmega_k}^2}^{F(\bx)},\\\nonumber \mathrm{subject\,to}\,& \ff_{\kf}(\bx^{\kf})=\bZero,\,\kf=0,...,K_\mathrm{f}-1,\\\nonumber& \bg_{\kg}(\bx^{\kg}) + \bq_{\kg}=\bZero,\,\kg=0,...,K_\mathrm{g}-1 
\end{eqnarray}
\begin{algorithm}[t]
	\caption{Augmented Lagrangian-Iterative Least Squares}\label{alg:our}
	\begin{algorithmic}
		\State $\left<\bOmega_k, \be_k(\cdot) \right>_{0:K-1}$, error factors
		\State $\left< \bblambda^k, \bP_\kf, \ff_\kf(\cdot) \right>_{0:K_\mathrm{f}-1}$, equality constraint factors
		\State $\left< \bbmu^k, \bP_\kg, \bg_\kg(\cdot) \right>_{0:K_\mathrm{g}-1}$, inequality constraint factors
		\State $I_{\mathrm{GN}}$, number of inner \gls{gn} iterations
		\While{!converged}		
		\For{$I_{\mathrm{GN}}$}
		\State $\bH \leftarrow \bZero$ 
		\State $\bb \leftarrow \bZero$
		\For{all factors}
		\State $\bb~\pluseq\bb^k,\,\bH~\pluseq\bH^k$ (\eqref{eq:linear-sys-unconstrained})
		\State $\bb~\pluseq\bb^\kf,\,\bH~\pluseq\bH^\kf$ (\eqref{eq:linear-sys-constrained})
		\State $\bb~\pluseq\bb^\kg,\,\bH~\pluseq\bH^\kg$ (\eqref{eq:linear-sys-constrained})
		\EndFor
		\State $\bDeltax\leftarrow \mathrm{solve}(\bH \bDeltax = -\bb)$
		\State $\hat \bx \leftarrow \hat\bx \boxplus \bDeltax$
		\EndFor
		\For{all constraint factors}
		\State $\bblambda^\kf \leftarrow \bblambda^\kf + 2\bP_{\kf}\ff_\kf(\bx^\kf)$ (\eqref{eq:dual-step})
		\State $\bbmu^\kg \leftarrow \max(\bZero, \bbmu^\kg + 2\bP_{\kg}\bg_\kg(\bx^\kg))$ (\eqref{eq:dual-step})
		\State Update $\rho$ (\eqref{eq:adaptive-rho})
		\EndFor
		\EndWhile
	\end{algorithmic}
\end{algorithm}

Hence, the Augmented Lagrangian for problem \eqref{eq:nlp} becomes
\begin{align}
	\label{eq:augmented-lagrangian}
&\mathcal{L}(\bx,\bq;\bblambda,\bbmu,\bP)=F(\bx)\\\nonumber&+\sum_{\kf=0}^{K_\mathrm{f}-1}\big[{\bblambda^{\kf}}^T\ff_{\kf}(\bx^{\kf})+|\ff_{\kf}(\bx^{\kf})||_{\bP_{\kf}}^2\big]\\\nonumber
	&+\sum_{\kg=0}^{K_\mathrm{g}-1}\big[{\bbmu^\kg}^T(\bg_\kg(\bx^\kg)+\bq_\kg)+||\bg_\kg(\bx^\kg)+\bq_\kg||^2_{\bP_{\kg}}\big]
\end{align}
Differently from~\cite{bertsekas2014ap}, we use here diagonal matrices $\bP_{\kf}=\mathrm{diag}(\rho_{\kf,1}, \rho_{\kf,2}, ..., \rho_{\kf,j})$ and $\bP_{\kg}=\mathrm{diag}(\rho_{\kg,1}, \rho_{\kg,2}, ..., \rho_{\kg,j})$, as big as the dimension of the constraints, instead of two scalar penalties $\rho_{\kf}$ and $\rho_{\kg}$. In this way, every component of the constraints is weighted by a different coefficient, which can be adapted based on the magnitude of the constraint violation along the corresponding dimension, rather than on the overall norm.

The Lagrangian method~\cite{bertsekas2014ap} iteratively minimizes 	\eqref{eq:augmented-lagrangian} with respect to $(\bx,\bq)$ for various values of $(\bblambda,\bbmu,\bP)$. 
If $\bx$ and $(\bblambda,\bbmu,\bP)$ are fixed, $\mathcal{L}(\bx,\bq;\bblambda,\bbmu,\bP)$ in \eqref{eq:augmented-lagrangian} can be minimized with respect to $\bq$. Furthermore, considering $\bP_\kg$ diagonal makes the minimization in each component $q_i$ of $\bq_\kg$ independent
\begin{equation}\label{eq:q-min}
q_i^*=\argmin_{{q_i}>0}{{\mu_{\kg,i}}(g_{\kg,i}(\bx^\kg)+{q_i})+\rho_{\kg,i}(g_{\kg, i}(\bx^\kg)+q_{i})^2}
\end{equation}
\eqref{eq:q-min} is a quadratic function in $q_{i}$, with unconstrained minimum $\hat{q}_{i}=-\left[\frac{\mu_{\kg,i}}{2\rho_{\kg,i}}+g_{\kg,i}(\bx^\kg)\right]$. Its global minimum subject to ${q_i}>0$ is therefore
\begin{equation}\label{eq:q-star}
	{q_i}^*=\max\left(0, -\left[\frac{\mu_{\kg,i}}{2\rho_{\kg,i}}+g_{\kg,i}(\bx^\kg)\right]\right)
\end{equation}

Let $\bg_{\kg}^+(\bx^\kg) = \bg_{\kg}(\bx^\kg)+\bq_{\kg}^*$, component-wise
\begin{equation}
g_{\kg,i}^+(\bx^\kg)=g_{\kg,i}(\bx^\kg)+q_i^*=\max\left(g_{\kg,i}(\bx^\kg), -\frac{\mu_{\kg,i}}{2\rho_{\kg,i}}\right)
\end{equation}
the Augmented Lagrangian for problem \eqref{eq:nlp} can be finally written as
\begin{align}
	\label{eq:al-active}
	&\mathcal{L}(\bx,\bq;\bblambda,\bbmu,\bP)=\sum_{k =0}^{K-1}||\be_k(\bx^k)||^2_{\bOmega_k}\\\nonumber&+\sum_{\kf=0}^{K_\mathrm{f}-1}\big[{\bblambda^{\kf}}^T\ff_{\kf}(\bx^{\kf})+||\ff_{\kf}(\bx^{\kf})||_{\bP_{\kf}}^2\big]\\\nonumber
	&+\sum_{\kg=0}^{K_\mathrm{g}-1}\big[{\bbmu^\kg}^T\bg_\kg^+(\bx^\kg)+||\bg_\kg^+(\bx^\kg)||^2_{\bP_{\kg}}\big]
\end{align}
Each term in parenthesis can be modeled as a factor in a factor-graph. In the next section, we specify how factors corresponding to constraints differ from regular error factors of classical \gls{ils} solvers.

\subsection{Augmented Lagrangian on Factor Graphs}
\label{sec:al-on-fg}
The \gls{al} method~\cite{bertsekas2014ap} is a primal-dual method for solving \eqref{eq:nlp} which computes the solution to the $\mathrm{maxmin}$ dual problem $\mathrm{max_{\bblambda, \bbmu} min_{\bx}} \mathcal{L}(\bx,\bblambda,\bbmu)$. At each iteration $i$, the primal step updates $\bx$ by minimizing  $\mathcal{L}(\bx,\bblambda,\bbmu)$ with fixed $(\bblambda,\bbmu)$. Our solver updates the current estimate of $\bx$ by taking $I_{\mathrm{GN}}$ Gauss-Newton steps over the \gls{al} function of~\eqref{eq:al-active}. 
The quadratic approximation of~\eqref{eq:al-active} is computed considering the first-order Taylor expansion of the error function and of the constraints around the current estimate $\hat{\bx}$ 
\begin{align}
		\be_k(\hat \bx^k \boxplus \bDelta \bx^k) &\simeq \hat{\be}_k + \bJ_k \bDelta\bx^k \\\nonumber
		\ff_\kf(\hat \bx^\kf \boxplus \bDelta \bx^\kf) &\simeq \hat{\ff}_\kf + \bF_\kf \bDelta\bx^\kf \\\nonumber
		\bg_\kg^+(\hat \bx^\kg \boxplus \bDelta \bx^\kg) &\simeq \hat{\bg}^+_\kg + \bG^+_\kg \bDelta\bx^\kg
\end{align}
Using the $\boxplus$ operator on the state manifold, the estimate is updated according to $\hat \bx \leftarrow \hat\bx \boxplus \bDeltax$, where $\bH^L \bDeltax = -\bb^L$:
\begin{align}
	\label{eq:linear-sys-constrained}
	\bb^L&=\bb+\sum_{\kf = 0}^{K_\mathrm{f}-1}\overbrace{\bF_\kf^T\bP_{\kf}\hat\ff_\kf\!+\!\frac{1}{2}{\bF_\kf}^T\bblambda^\kf}^{\bb^\kf}\\\nonumber&+\sum_{\kg=0}^{K_\mathrm{g}-1}\overbrace{{\bG^+_\kg}^T\bP_{\kg}\hat\bg^+_\kg\!+\!\frac{1}{2}{\bG^+_\kg}^T\bmu^\kg}^{\bb^\kg}
	\\\nonumber
	\bH^L&=\bH+\sum_{\kf=0}^{K_\mathrm{f}-1}\overbrace{\bF_\kf^T\bP_{\kf}\bF_\kf}^{\bH^\kf}+\sum_{\kg = 0}^{K_\mathrm{g}-1}\overbrace{{\bG^+_\kg}^T\bP_{\kg}\bG^+_\kg}^{\bH^\kg}.
\end{align}
with $\bH$ and $\bb$ from \eqref{eq:linear-sys-unconstrained}. Hence, $\kf$-th and $\kg$-th constraint factors contribute to $\bH^L$ and $\bb^L$, respectively with $\bH^{\kf}, \bb^\kf$ and $\bH^{\kg}, \bb^\kg$. 

The dual step happens within the constraint factors where $(\bblambda, \bbmu)$ are updated by taking projected gradient ascent step over $\mathcal{L}(\bx,\bblambda,\bbmu)$ with fixed $\bx$ weighted by the penalty coefficients $\bP_{\ff}, \bP_{\bg}$
\begin{equation}
	\label{eq:dual-step}
	\begin{cases}
		\bblambda^\kf \leftarrow \bblambda^\kf + 2\,\bP_{\kf}\ff_\kf(\bx^\kf),\,\kf = 0,...,{K_\mathrm{f}-1}\\
		\bbmu^\kg \leftarrow \max(\bZero, \bbmu^\kg + 2\,\bP_{\kg}\bg_\kg(\bx^\kg)),\,\kg = 0,...,{K_\mathrm{g}-1}
	\end{cases}
\end{equation} 
Penalty parameters in $\bP_{\kf}$ and $\bP_{\kg}$ are usually adapted based on the evolution of the constraint violation between subsequent iterations~\cite{sodhi2020icra}. 

In the following, we present the adaptation scheme we used. Let us denote as $\rho$ the coefficient associated to the constraint $\ff=0$, as $\ff^{-} = \mathrm{max}(0, |\ff_{i-1}|-|\ff_{i}|/|\ff_{i-1}|)$ the percentage decrease in constraint violation from iteration $i-1$ to iteration $i$, and as $\ff^{+} = \mathrm{max}(0, (|\ff_i| - |\ff_{i-1}|)/|\ff_i|)$ the percentage increase in constraint violation. At iteration $i$, 
our choice is to compute $\rho$ in the range $(\rho_\mathrm{m}, \bar{\rho})$ if $\ff^{+}$ is positive, or $(\bar{\rho}, \rho_\mathrm{M})$ if $\ff^{-}$ is positive
\begin{equation}
\begin{cases}
	\label{eq:adaptive-rho}
	\rho = \bar{\rho}_i + \ff^{-}(\rho_{\mathrm{M}}-\bar{\rho}_i) + \ff^{+}(\rho_{\mathrm{m}}-\bar{\rho}_i)\\
	\bar{\rho}_{i+1} = \bar{\rho}_i + \ff^{-}(\rho_{\mathrm{M}}-\bar{\rho}_i).
\end{cases}
\end{equation}
$\bar{\rho}$ changes over the iterations to guarantee that the increase in $\rho$ due to reduction in constraint violation is kept across subsequent iterations. Otherwise, constant constraint violation would result in decreasing $\rho$. Clamping $\rho$ within the interval $(\rho_\mathrm{m},\rho_\mathrm{M})$ prevents the algorithm from diverging in case of bad initial guesses, while allowing larger values to be used when constraint satisfaction is improving. In all practical applications described in the remainder, we use $\bar{\rho}_0 = 1.0, \rho_{\mathrm{m}}=0.5, \rho_{\mathrm{M}}=2.0$. Further, all Lagrange Multipliers are initialized at zero in the applications.

\section{Applications}
\label{sec:exp}
The objective of this work is to present a methodology to address general constrained optimization problems using factor graphs. In the following we present three applications to show the capabilities of our approach: 
(i) improve in performance thanks to inclusion of constraints in pose estimation; (ii) alternative approach to rotation synchronization which directly includes rotation matrix constraints; (iii) runtime advantage compared to \gls{ipopt} in \gls{mpc}.

\subsection{Constrained Pose Estimation}
\label{sec:est}

\begin{figure}[t!]
	\centering
	\includegraphics[width=0.99\linewidth]{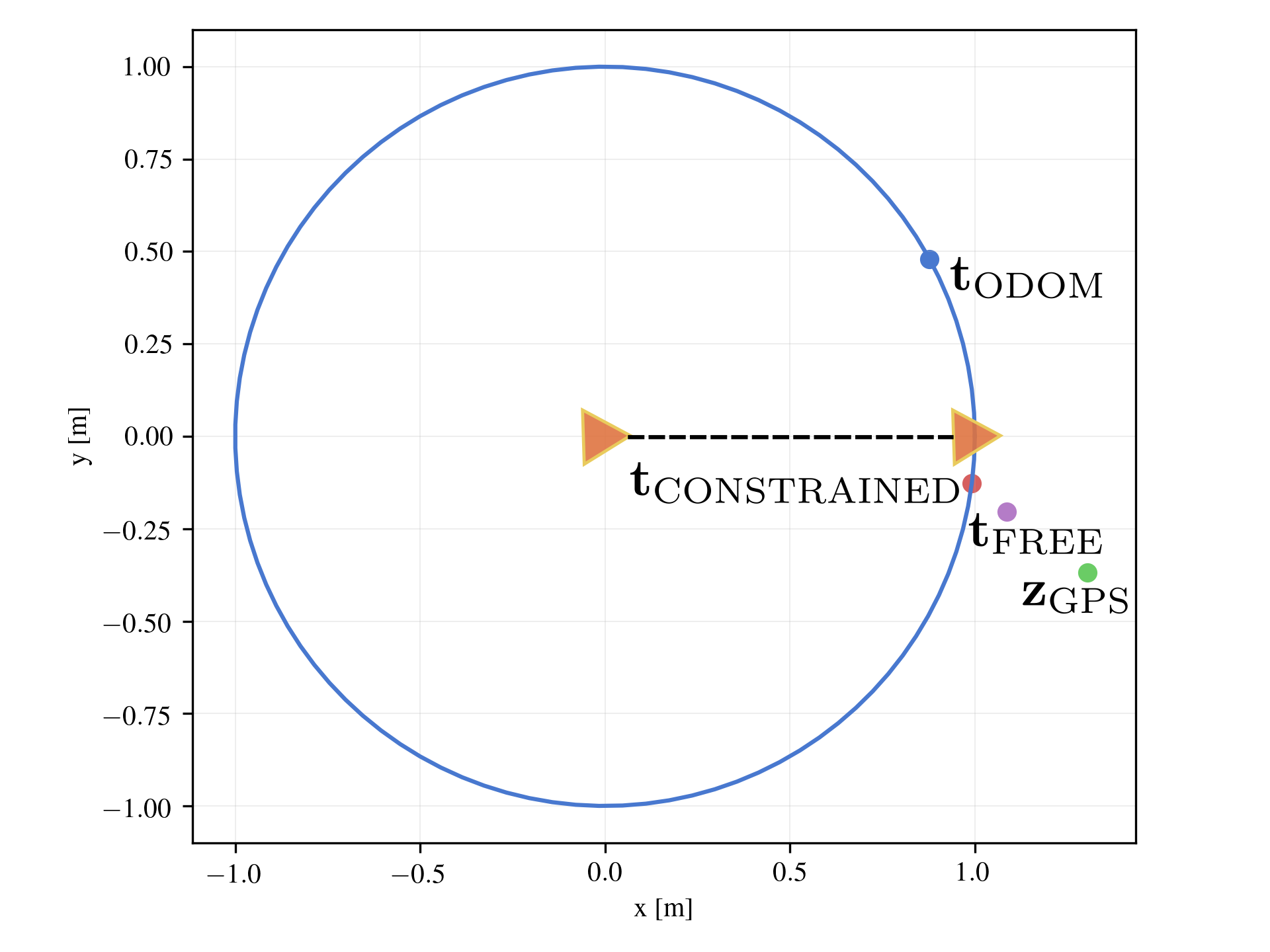}
	\caption{Representation of the constrained estimation problem. By adding the constraint~\eqref{eq:circumference-constraint} to the optimization problem, $\bt_{\mathrm{CONSTRAINED}}$ is closer to Ground Truth $[1,0,0]$ compared to $\bt_{\mathrm{FREE}}$.}
	\label{fig:estimation-plt}
	\vspace{-10px}
\end{figure}

2D pose estimation is the problem of determining robot position and heading $\bX^W_R=[\bt;\,\theta]=[x,\,y,\,\theta]\in\mathrm{SE}(2)$ that maximize the likelihood of the measurements. As an illustrative example of how the capacity of handling constraints can improve the performance, we address here the 2D navigation application of Barrau~\etal\cite{barrau2020tac}. A unicycle starts from perfectly known position $\bt=[0,0]^T$ with unknown heading. It drives in straight line with constant linear velocity $v$ and zero angular velocity $\omega$ for known time $\mathrm{T}$. It then receives a GPS measurement of its new position $\bz_{\mathrm{GPS}}\in\mathbb{R}^2$, and uses it to correct the odometry measurement $\bZ_{\mathrm{ODOM}}=[\bt_{\mathrm{ODOM}};\,\theta_{\mathrm{ODOM}}]\in\mathrm{SE}(2)$, obtained integrating the unicycle kinematics from the initial guess $\bX^W_{R, \mathrm{ig}}=[0,0,\theta_0]$. \figref{fig:estimation-plt} illustrates the problem assuming $v=1\mathrm{m/s}$, $\mathrm{T}=1\mathrm{s}$ and zero ground-truth orientation (orange triangles).

\begin{figure}[t]
	\centering
	\includegraphics[width=0.4\linewidth]{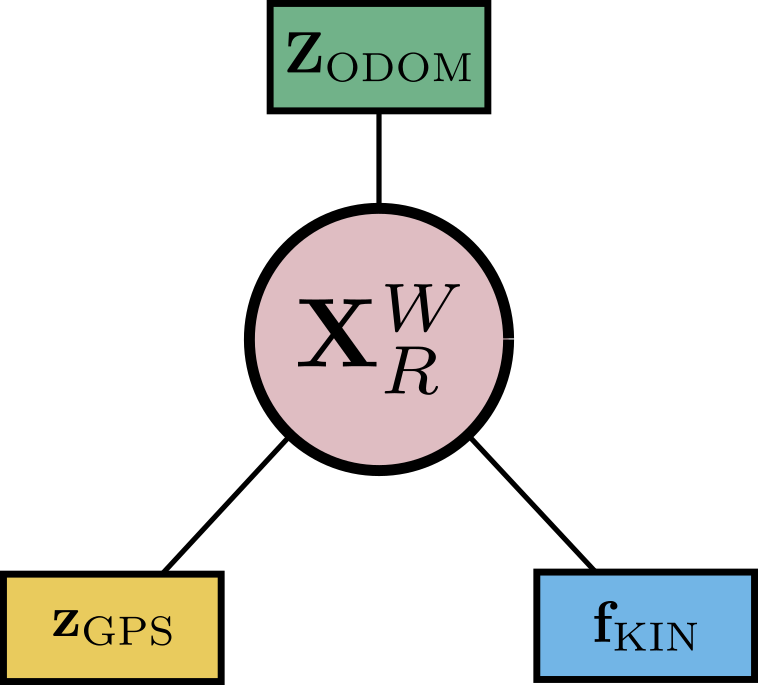}
	\caption{Factor graph modeling the pose estimation problem subject to kinematics which constrains the pose on a circumference with radial orientation.}
	\label{fig:estimation-factor-graph}
\end{figure}

\begin{figure}[t]
	\centering
	\includegraphics[width=0.99\linewidth]{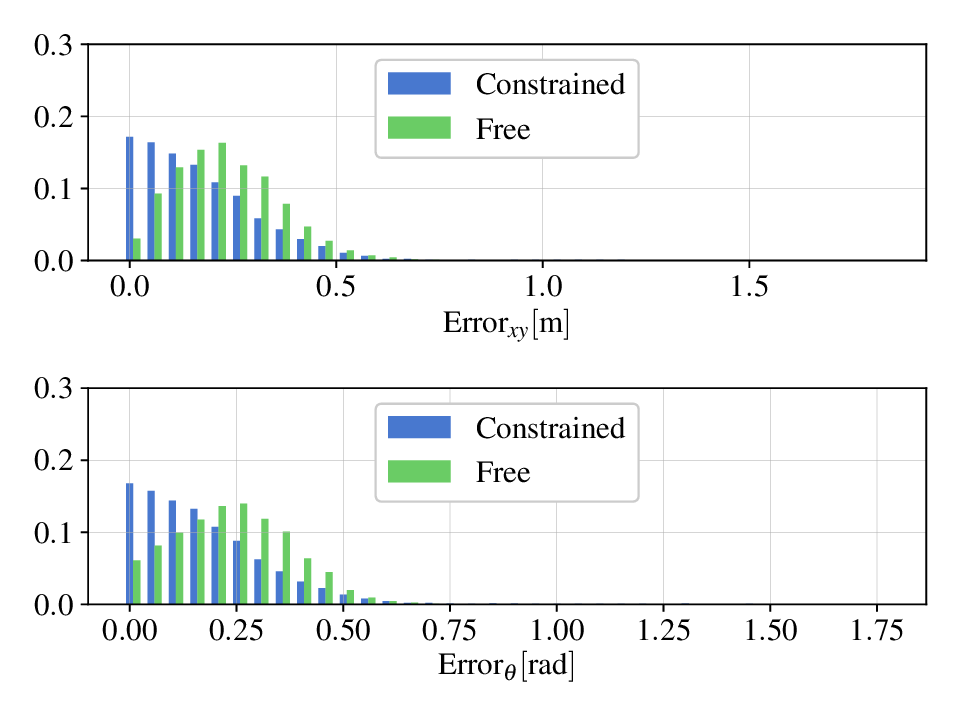}
	\caption{Probability distribution of linear and rotational error of the estimate with and without the constraint of~\eqref{eq:circumference-constraint}. Each test corresponds to a different $\bz_{\mathrm{GPS}}$ extracted from the normal distribution $\mathcal{N}([1,0,0]; \bOmega_{\mathrm{GPS}})$, with $\theta_0 = 0.5\mathrm{rad}$.}
	\label{fig:estimation-error}
\end{figure}

Traditionally, the estimate of the robot pose $\bX^W_R$ is obtained by solving
\begin{eqnarray}
	\label{eq:constrained-estimation}
	F(\bX^W_R)=||\bt-\bz_{\mathrm{GPS}}||_{\bOmega_{\mathrm{GPS}}}^2+||{\bX^W_R}^{-1}\bZ_{\mathrm{ODOM}}||_{\bOmega_{\mathrm{ODOM}}}^2
\end{eqnarray}
It finds the estimate which best explains both odometry and GPS measurements. 
However, it neglects the information that the robot is moving on a straight line, which implies (a) robot position on the circumference (b) the heading $[\cos(\theta), \sin(\theta)]^T$ is radial. The two conditions are summarized by
\begin{equation}
\label{eq:circumference-constraint}
\ff_{\mathrm{KIN}}(\bX)=
\begin{pmatrix}
x^2+y^2-(v\mathrm{T})^2\\
x\sin(\theta)-y\cos(\theta)
\end{pmatrix}=
\begin{pmatrix}
0\\
0
\end{pmatrix}
\end{equation}
The constraint factor modeling~\eqref{eq:circumference-constraint} is represented by the blue square in \figref{fig:estimation-factor-graph}. As any other factor, it is connected to the variable on which it depends.  
\figref{fig:estimation-error} shows the probability distribution of the translational and rotational error obtained over 10K experiments with $\bOmega_{\mathrm{ODOM}}=10\,\bI_{3\times3}$ and $\bOmega_{\mathrm{GPS}}=20\,\bI_{3\times3}$.  Lower errors are more likely imposing~\eqref{eq:circumference-constraint}.

\subsection{Rotation Synchronization}
The second application we consider is Rotation Synchronization in $\mathbb{R}^3$~\cite{eriksson2021tpa}. It is the $\so3$ instance of the Group Synchronization problem which consists in finding the elements of a group, in our case $\bR_i\in\mathbb{R}^{3\times3}$ with $\bR_i^T\bR_i=\bI_{3\times3};\,\det\bR_i=1$, starting from pairwise measurements $\bZ^i_j$, in our case $\bR_i^T\bR_j$. It is a sub-problem of many applications from \gls{sfm} to pose graph optimization. In absence of a good initial guess, a solution is obtained by finding a set of matrices $\bA\in\mathbb{R}^{3\times3}$ that minimize
\begin{equation}
	\label{eq:rot-sync}
	F(\bA_{0:n})=\sum_{i, j}||\mathrm{flatten}(\bA_i\bZ^i_{j}-\bA_j)||_{\bOmega_{i,j}}^2
\end{equation}
where the operator $\mathrm{flatten}(\cdot):\mathbb{R}^{3\times3}\leftarrow\mathbb{R}^9$ stacks the rows of the input matrix into a vector. Once a solution is found, the closest rotation matrices $\bR^*_{0:n}$ are obtained by \gls{svd}
\begin{equation}
\bR^*_i = \bU_i^T\bV_i\leftarrow\bU_i, \bS_i, \bV_i =\mathrm{SVD}(\bA_i)
\end{equation}

Using our framework instead, we can embed the rotation constraints 
\begin{equation}
	\label{eq:orthogonality-constraint}
	\ff(\bA_i)= 
	\begin{pmatrix}
	\mathrm{flatten}(\bA_i^T\bA_i-\bI_{3\times3})\\
	\det(\bA_i) -1
	\end{pmatrix}=
\begin{pmatrix}
	\bzero_9\\
	0
\end{pmatrix}
\end{equation}
directly in the factor graph so that the solution of \eqref{eq:rot-sync} subject to \eqref{eq:orthogonality-constraint} provides valid rotation matrices by construction. Constraint factors modeling~\eqref{eq:orthogonality-constraint} are represented by blue squares.

\begin{figure}[t]
	\centering
	\includegraphics[width=0.6\linewidth]{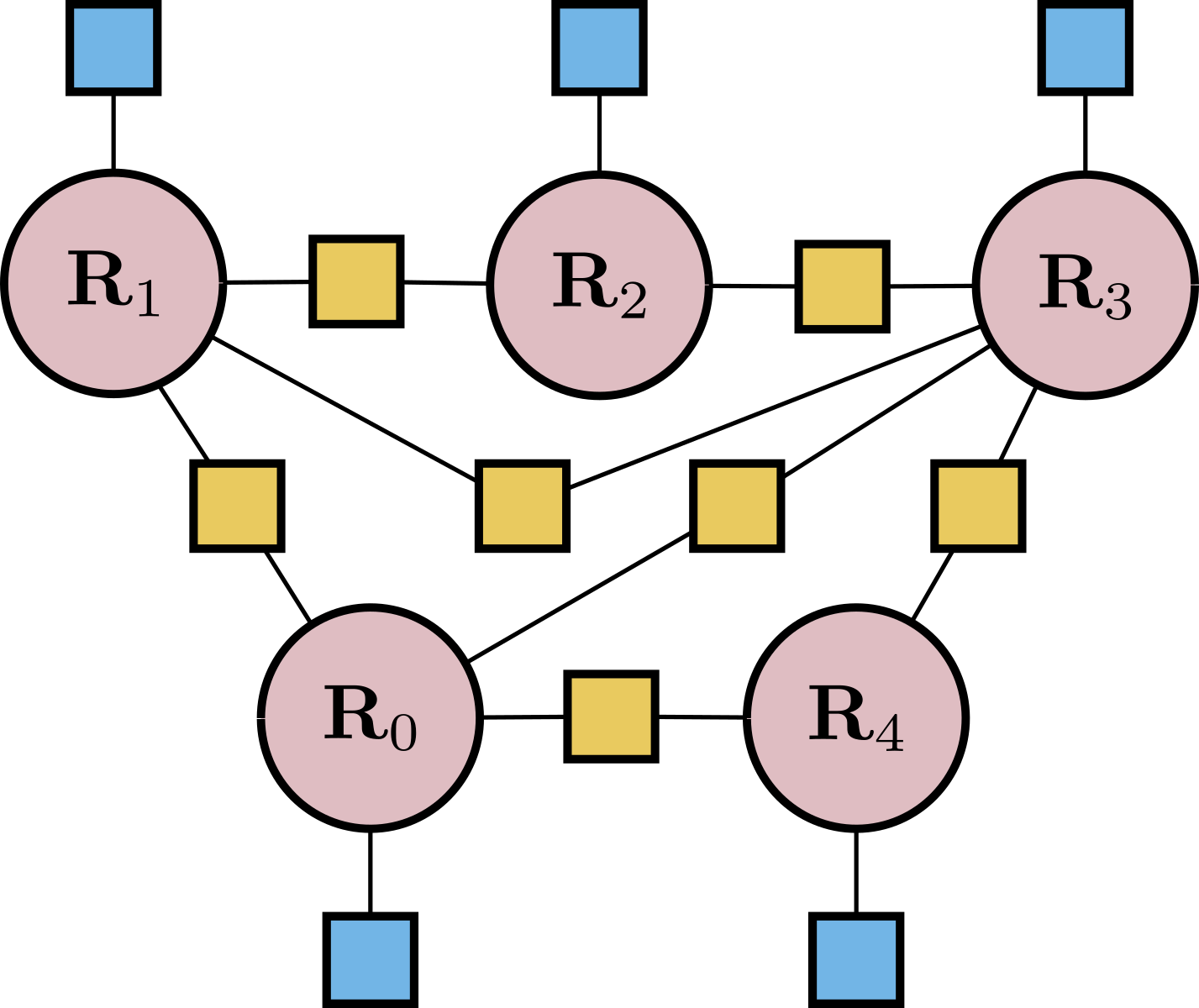}
	\caption{Factor graph modeling the rotation synchronization problem, $n=4$.}
	\vspace{-5px}
	\label{fig:sfm-factor-graph}
\end{figure}

\begin{table}[t]
	\centering
	\caption{Mean angular error [$\mathrm{rad}$] compared to Ground Truth, $n=99$ }
	\label{tab:sfm-error}
	\begin{subtable}{0.45\textwidth}
		\caption{$\Delta\alpha_x$}
		\begin{center}
			\vspace{-10px}
			\begin{tabular}{ |c|c|c|c| } 
				\hline
				$||\bOmega||_{\infty}$ & Constrained& SVD+Quaternion & SVD\\
				\hline
				1$\me$3  & 7.792$\me$-5  & 7.792$\me$-5& 7.818$\me$-5\\
				5$\me$3 & 6.448$\me$-6  & 6.448$\me$-6 & 6.701$\me$-6\\
				1$\me$4 & 2.978$\me$-6 & 2.977$\me$-6 & 3.011$\me$-6 \\
				\hline 
			\end{tabular}
		\end{center}
	\end{subtable}
	\begin{subtable}{0.45\textwidth}
		\vspace{10px}
		\caption{$\Delta\alpha_y$}
		\vspace{-10px}
		\begin{center}
			\begin{tabular}{ |c|c|c|c| } 
				\hline
				$||\bOmega||_{\infty}$ & Constrained& SVD+Quaternion & SVD\\
				\hline
				1$\me$3 & 1.038$\me$-4  & 1.038$\me$-4& 1.038$\me$-4\\
				5$\me$3 & 7.664$\me$-6 & 7.665$\me$-6 & 8.011$\me$-6\\
				1$\me$4 & 2.382$\me$-6 & 2.381$\me$-6& 2.469$\me$-6 \\
				\hline
			\end{tabular}
		\end{center}
	\end{subtable}
	\begin{subtable}{0.45\textwidth}
		\vspace{10px}
		\caption{$\Delta\alpha_z$}
		\vspace{-10px}
		\begin{center}
			\begin{tabular}{ |c|c|c|c| } 
				\hline
				$||\bOmega||_{\infty}$ & Constrained& SVD+Quaternion & SVD\\
				\hline
				1$\me$3  & 1.074$\me$-4 & 1.074$\me$-4 & 1.067$\me$-4\\ 
				5$\me$3 & 1.229$\me$-5 & 1.229$\me$-5& 1.234$\me$-5\\			
				1$\me$4 & 2.925$\me$-6 & 2.924$\me$-6& 2.923$\me$-6 \\			
				\hline
			\end{tabular}
		\end{center}
	\end{subtable}
\end{table}

\tabref{tab:sfm-error} compares our constrained approach with (a) traditional \gls{svd} method; (b) cascade of (a) and nonlinear quaternion synchronization. Average estimation errors are similar in all the three components $\alpha_x, \alpha_y, \alpha_z$. We show results for various values of the information matrix of the measurements. The same mechanism can be straightforwardly extended to synchronization problems in the special euclidean group $[\bR, \bt]\in\se3$, or in the similarity group $[s\bR, \bt]\in\sim3$, with $s\in\mathbb{R}$ scaling scalar factor.

\begin{figure}[t]
	\centering
	\includegraphics[width=0.99\columnwidth]{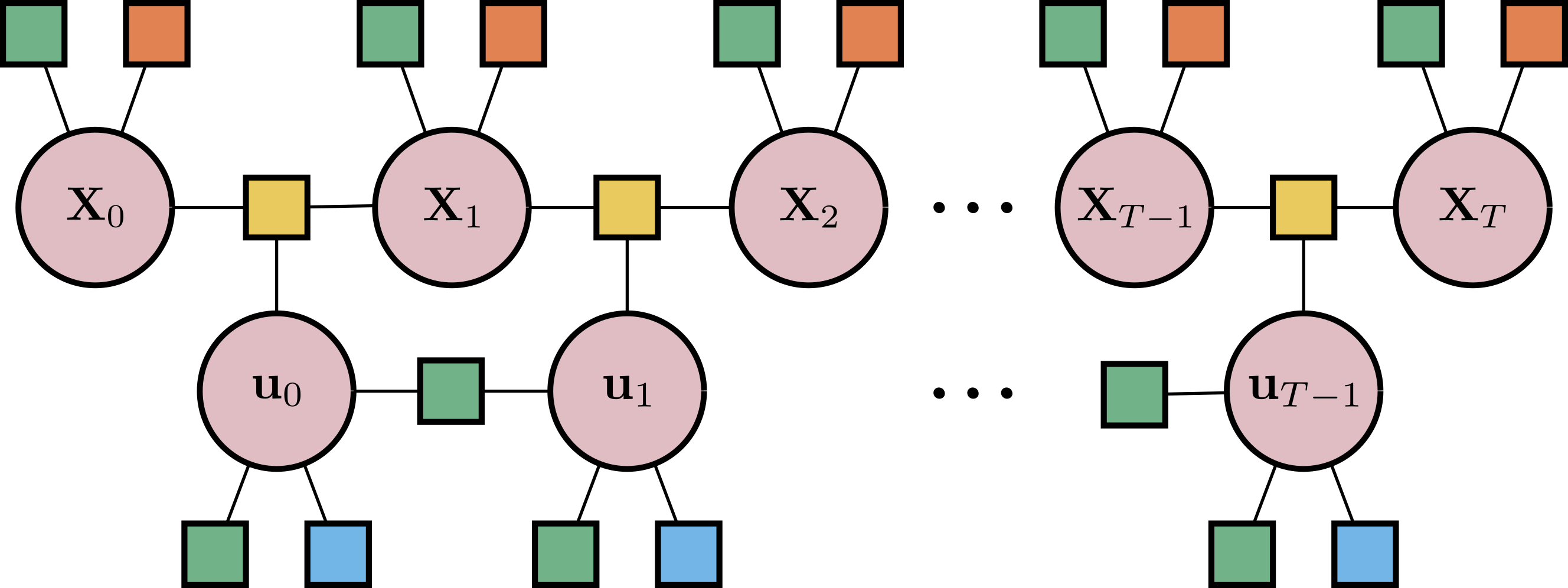}
	\caption{Factor graph modeling the nonlinear \gls{mpc} control problem.}
	\label{fig:factor-graph}
\end{figure}
\begin{figure}[t]
	\centering
	\includegraphics[width=0.55\linewidth]{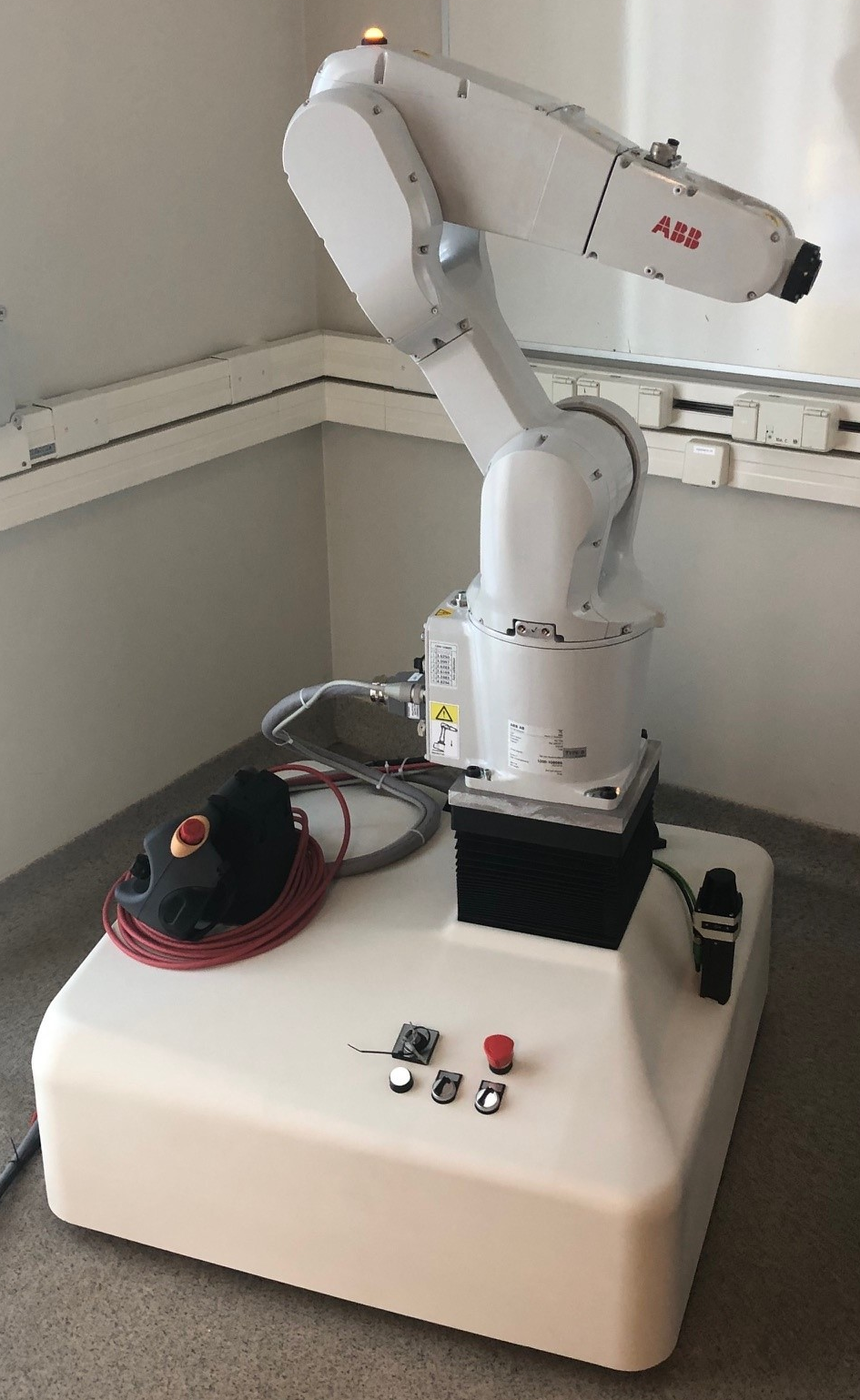}
	\caption{Our pseudo-omnidirectional platform with four steering and driving wheels.}
	\label{fig:roboclean}
\end{figure}
\begin{figure}[t]
	\begin{center}
		\begin{tabular}{cc}
			\includegraphics[width=0.45\columnwidth]{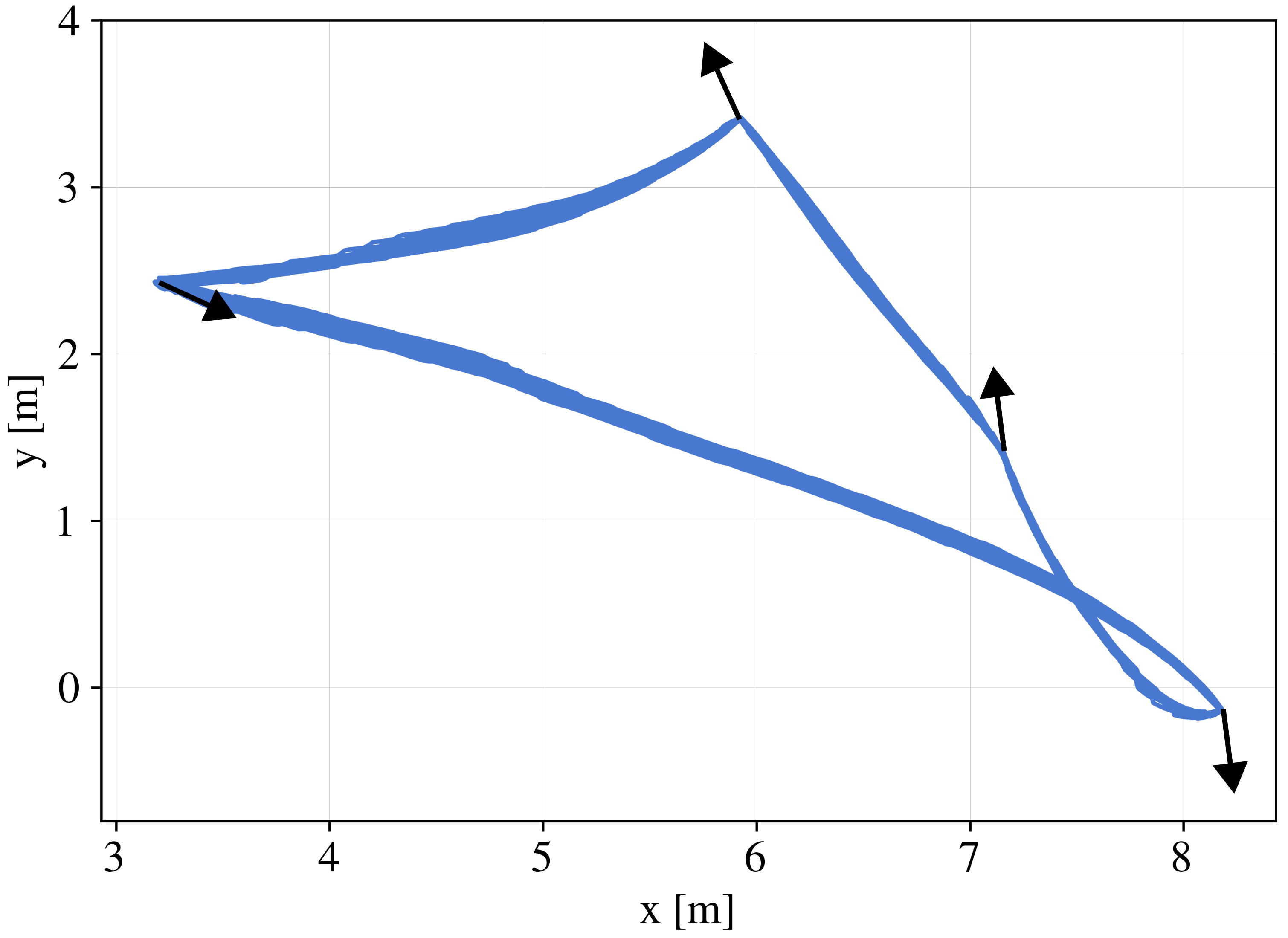}
			&
			\includegraphics[width=0.45\columnwidth]{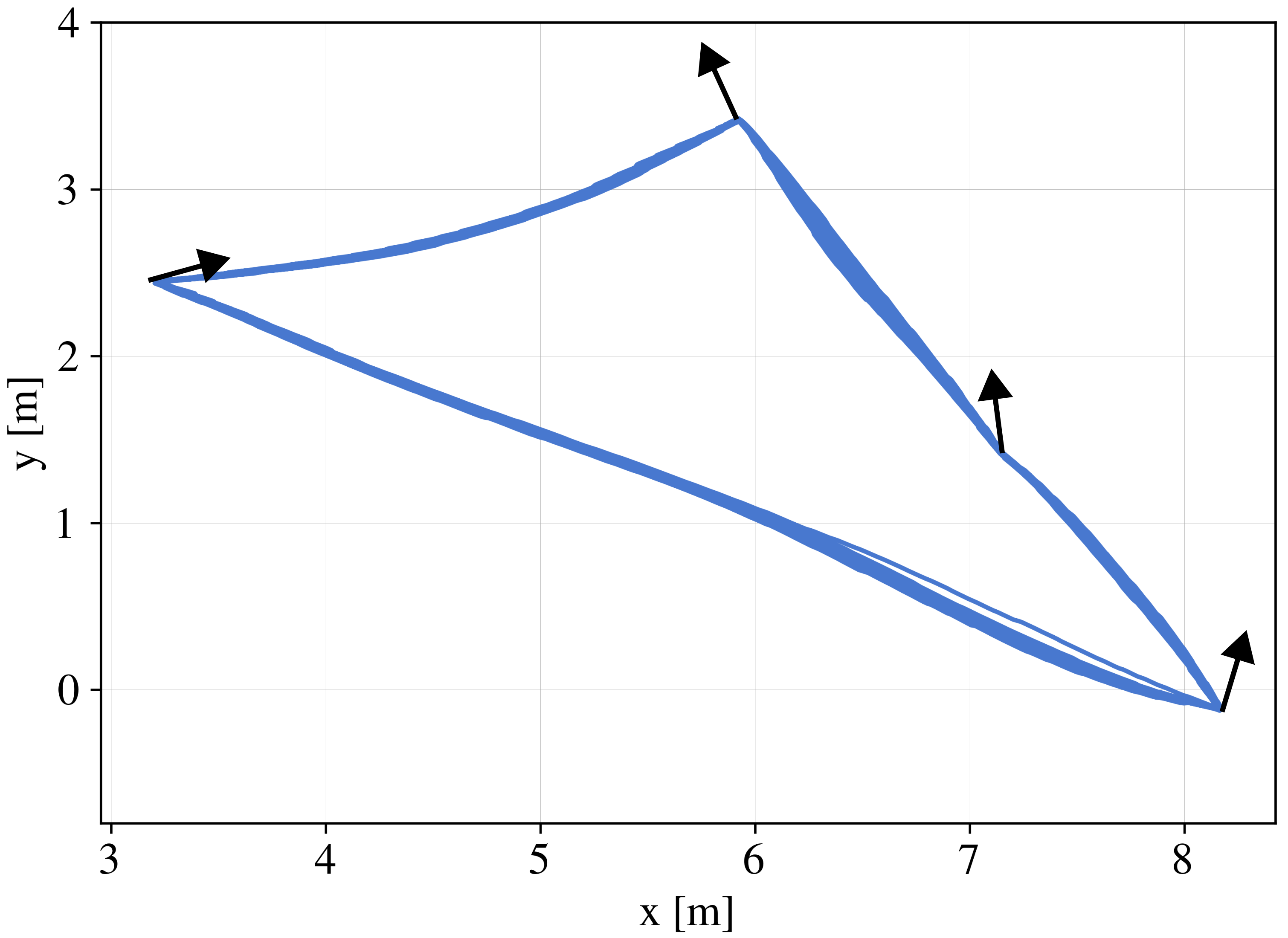}
		\end{tabular}
		\caption{Two sample set of goals, grouped in closed paths. The desired orientation of the goal is represented by the arrows.}
		\label{fig:trajectories}
	\end{center}
\end{figure}

\begin{figure}[t]
	\centering
	\includegraphics[width=0.8\columnwidth]{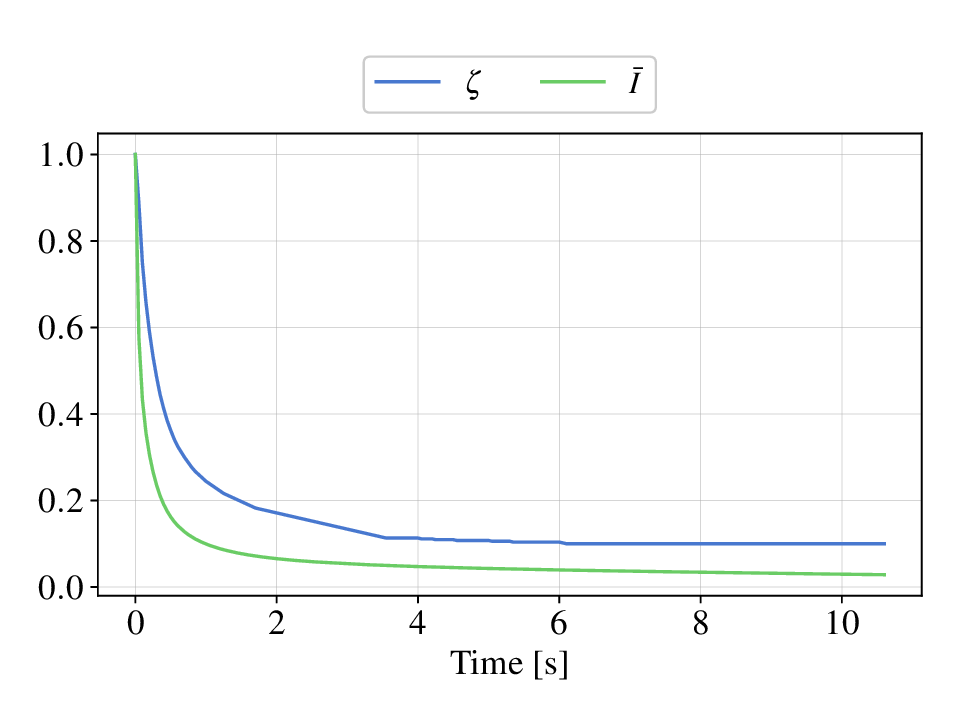}
	\caption{Evolution of damping $\zeta$ and of average number of iterations $\bar{I}$ on one round from goal 5 to goal 6. $\bar{I}$ is divided by its maximum value (1000).}
	\label{fig:damping}
\end{figure}

\subsection{Factor Graph-\gls{mpc}}
The pseudo-omnidirectional platform we used for experimental validation of our factor graph-\gls{mpc} is shown in \figref{fig:roboclean}. 
 The robot lives in a 2D world. Let us indicate by $\bx = [x,y,\theta]^T$ its position and orientation, $v$ the module of the linear velocity, $\phi$ its direction, and $\omega$ the angular velocity along the yaw axis. The kinematics is modeled by 
\begin{equation}
	\label{eq:kinematics}
\dot \bx
=\begin{pmatrix}
\bR(\theta)&\bzero_{2\times1}\\
\bzero_{1\times2}& 1
\end{pmatrix}\begin{pmatrix}
\cos(\phi)&0\\
\sin(\phi)&0\\
0&1
\end{pmatrix}
\begin{pmatrix}
v\\\omega
\end{pmatrix}
\end{equation}
The robot is controlled in acceleration by $\bu=[dv, d{\phi}, d{\omega}]^T$.  At each \gls{mpc} step,  the following objective function is minimized to get the robot reach the goal $\bg\in\mathbb{R}=[g_x,g_y,g_\theta]^T$
\begin{align}
	\label{eq:J}
	J(\bx) &= \sum_{t=0}^{T}\big|\big|\bg-\bx_t\big|\big|_{\bOmega^\mathrm{g}_t}^2\\\nonumber
	&+ \sum_{t=0}^{T-1} ||\bu_t||_{\bOmega^\mathrm{u}_t}^2+ \sum_{t=0}^{T-2} ||\bu_{t+1} - \bu_t||_{\bOmega^{\mathrm{du}}_t}^2
\end{align}
where $t$ indicates the time instants of the \gls{mpc} horizon, which has length $T$. Each term in \eqref{eq:J} corresponds to a regular error factor and is shown with a green square. Compared to our reference~\cite{andreasson2022mdpi}, we added the third summation, which penalizes changes between consecutive acceleration inputs to reduce the jerk. 

The variables of the factor graph modeling the optimal control problem are: the states of the robot $\bX_{t=0:T}=[x_t,y_t,\theta_t,v_t,\phi_t,\omega_t]^T$ and the controls $\bu_{t=0:T-1}$. Indeed, the solution of our factor graph solver is the optimal set of joint states and controls. State $\bX_{t}$ and control $\bu_{t}$ are linked to state $\bX_{t+1}$ at time $t+1$ by the nonlinear dynamics constraint $\bX_{t+1} = \bF(\bX_t,\bu_t)$, resulting from Runge-Kutta RK4 integration of \eqref{eq:kinematics}, jointly with $\dot{v}=dv, \dot{\phi}=d\phi, \dot{\omega}=d\omega$. The nonlinear constraint becomes a factor of degree three shown by the yellow squares in \figref{fig:factor-graph}. The velocity components of the states $\bX=\bX_{0:N-1}$ are subject to the following constraints:
\begin{equation}
	\label{eq:maximum-velocity}
	\left|\omega - \frac{v}{d}\right|\leq {\omega}^{\mathrm{max}},\qquad
	\left|\omega + \frac{v}{d}\right|\leq {\omega}^{\mathrm{max}}.
\end{equation}
These constraints are illustrated by orange squares in the figure.
Similarly the inputs $\bu=\bu_{0:N-1}$ should satisfy the following acceleration limits:
 \begin{equation}
	       \label{eq:maximum-acceleration}
	       \left|dv\right|\leq dv^{\mathrm{max}},\qquad
	       \left|d\phi\right|\leq d\phi^{\mathrm{max}},\qquad
	       \left|d\omega\right|\leq d\omega^{\mathrm{max}},
\end{equation}
that are the blue boxes in \figref{fig:factor-graph}.

When a new potentially far goal is set, this iterative schema spends substantial effort if the initial guess is poor. Whereas in a practical application providing a reasonable initial guess would dramatically enhance the performances, we choose to highlight the behavior of our system by initializing all poses in the origin, and all velocities and controls to zero each time a new goal is set.

An analysis of the linear system of  \eqref{eq:linear-sys-constrained} under this poor initial guess, reveals that it is under-constrained. However as the solution becomes closer to the optimum the system becomes better conditioned. Therefore, we add a damping term to the linear system of the primal update, which becomes $(\bH^L + \zeta\bI)\bDelta\bx = -\bb^L$. Using a high value of $\zeta$ would address the ill-conditioning issue, but the solution requires more iterations and can be a local minimum. Hence, at each epoch we use the previous solution as initial guess and select the $\zeta$ based on its quality. 

Our approach is to use an adaptive scheme on $\zeta$ which is kept constant within one MPC epoch, but whose value depends on the status of the solver at the previous epoch.
In general the  higher the number of iterations the more complex the problem is.
This intuition can be translated onto adaptive scheme to modulate $\zeta$ between epochs, depending on the history of iterations.
Our straightforward choice is to let $\zeta$ vary in the range $(\zeta_\mathrm{m}, \zeta_\mathrm{M})$,
based on the average number of iterations $\bar{I}$ between all epochs as follows.
\begin{equation}
	\zeta=
	\begin{cases}
		\zeta_\mathrm{m},\, \mathrm{if}\,\bar I \leq \bar I_\mathrm{m}\\
		\zeta_\mathrm{m} + \frac{\zeta_\mathrm{M}-\zeta_\mathrm{m}}{\bar I_\mathrm{M}-\bar I_\mathrm{m}}*(\bar I-\bar I_\mathrm{m}),\, \mathrm{if}\,\bar I_\mathrm{m} < \bar I \leq \bar I_\mathrm{M}\\
		\zeta_\mathrm{M},\, \mathrm{if}\,\bar I > \bar I_\mathrm{M}.\\
	\end{cases}
	\label{eq:adaptive-zeta}
\end{equation}
In \eqref{eq:adaptive-zeta} we clamp the value of $\bar{I}$ in the range 
$(\bar I_\mathrm{m}, \bar I_\mathrm{M})$.
In the experiments we set $\bar I_\mathrm{m}=20, \bar I_\mathrm{M}=500$ and $\zeta_\mathrm{m} = 0.1,\zeta_\mathrm{M}=1.0$. \figref{fig:damping} shows the evolution of $\zeta$ and $\bar I$ as the epochs evolve. 
As a termination criterion for our solver we use $||\bDeltax||_{2}<\epsilon_{\bx}$, $||\ff_\kf(\bx)||_{\infty}<\epsilon_{\ff},\,\forall\kf=0:K_{\mathrm{f}}-1$, and $||\mathrm{max}(\bg_\kg, \bZero)||_{\infty}<\epsilon_{\bg},\,\forall\kg=0:K_{\mathrm{g}}-1$, with $\bZero\in\mathbb{R}^{\kg,J}$, and $\epsilon_{\bx}, \epsilon_{\ff}, \epsilon_{\bg} = 1\mathrm{e}{-3}$. The maximum allowed number of iterations is 1000. 

We carried real world experiments on the real robot equipped with an onboard computer Intel NUC I7 to validate the approach.
Subsequently, to gather a statistically significant measure of the performances, we constructed a simulated environment using ROS and Gazebo reflecting the real scenario, and we confirmed that the behaviors were equivalent in the two cases. In the latter case, we used a laptop Intel(R) Core(TM) i7-10750H CPU running at 2.60GHz with 16GB of RAM. 
We then  instructed the robot in the simulator to travel across 11 goals grouped in 3 closed paths, shown in \figref{fig:motivation}-(c) and \figref{fig:trajectories}. We repeated the experiment 30 times comparing the IPOPT solver used by Andreasson \etal~\cite{andreasson2022mdpi}, and our factor graph solver with two different Lagrangian functions: the one in \eqref{eq:al-active} and the one in \eqref{eq:al-qadri} introduced in~\cite{qadri2022incopt}. 

\begin{figure}[t]
	\centering
	\includegraphics[width=0.85\linewidth]{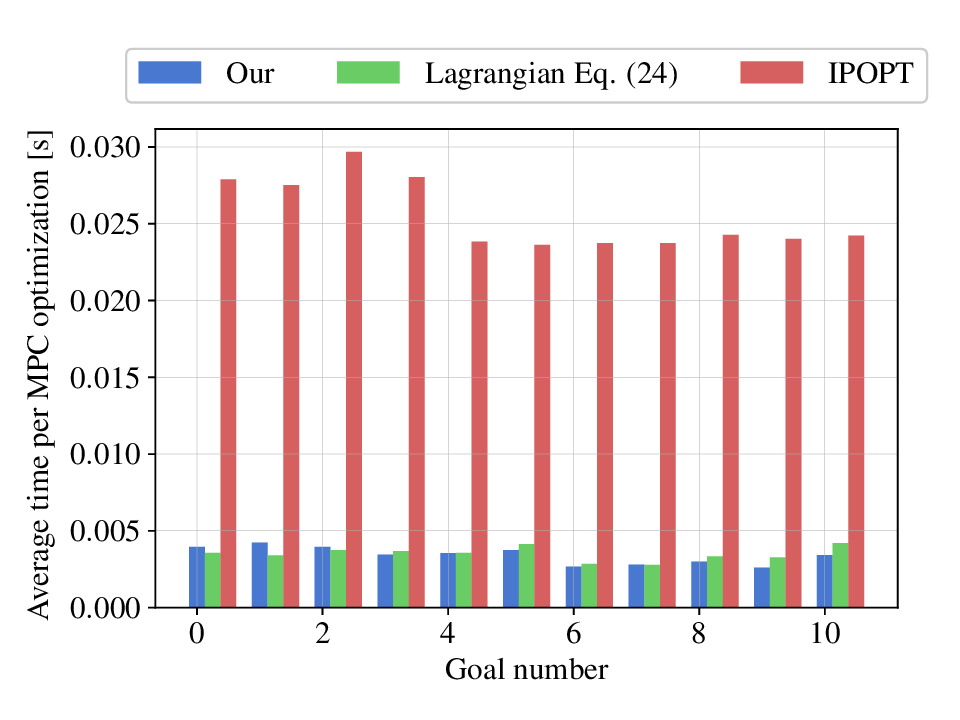}
	\caption{Comparison of average runtime per MPC optimization.}
	\label{fig:runtime-avg}
\end{figure}

Instead of~\eqref{eq:al-active}, in \cite{qadri2022incopt} they use as \gls{al} function
\begin{align}
	\mathcal{L}&(\bx;\bblambda,\bbmu)=F(\bx)
	+\sum_{\kf=0}^{K_\mathrm{f}-1}\big[{\bblambda^\kf}^T\ff_\kf(\bx^\kf)+||\ff_\kf(\bx^\kf)||_{\bP_{\kf}}^2\big]\\\nonumber		&+\sum_{\kg=0}^{K_\mathrm{g}-1}\big[{\bbmu^\kg}^T\bg_\kg(\bx^\kg)+||\mathrm{max}(\mathbf{0},\bg_\kg(\bx^\kg))||^2_{\bP_{\kg}}\big]
	\label{eq:al-qadri}	
\end{align}
To capture this Lagrangian, we modify $\bH^\kg$ and $\bb^\kg$ of all inequality constraint factors.
Let us indicate with $\hat\bg_\kg^+=\max(\bzero, \bg_\kg(\hat\bx^\kg))$, $\bG_\kg=\frac{\partial\bg_\kg(\bx\boxplus\Delta\bx)}{\partial\bDelta\bx}\big|_{\Delta\bx=\bzero}$ and $\bG^+_\kg=\frac{\partial\max(\bzero, \bg_\kg(\bx\boxplus\Delta\bx))}{\partial\bDelta\bx}\big|_{\Delta\bx=\bzero}$.
Instead of the terms in~\eqref{eq:linear-sys-constrained}, we get
\begin{align}
&\bb^{\kg}={\bG^+_\kg}^T\bP_{\kg}\hat\bg^+_{\kg}+\frac{1}{2}\bG_\kg^T\bbmu^{\kg}
\\\nonumber
&\bH^\kg={\bG^+_\kg}^T\bP_{\kg}\bG^+_\kg
\end{align}
The dual update follows the same rule of~\eqref{eq:dual-step}
\begin{equation}
\bbmu^\kg \leftarrow \max(\bZero, \bbmu^\kg + 2\,\bP_{\kg}{\bg}_\kg(\bx^\kg)),\,\kg = 0,...,{K_\mathrm{g}-1}
\end{equation}
Using the same \gls{ils} solver allows to isolate the effect of the different implementations of the constraint factors in the \gls{mpc} problem.
\figref{fig:runtime-avg} shows the average time per MPC optimization. Our approach is around 7 times faster than using \gls{ipopt}, as it can be seen in the figure. The difference with our implementation of \cite{qadri2022incopt} is less evident because the only difference is the choice of the \gls{al} function. 

\begin{figure}[t]
	\centering
	\includegraphics[width=0.85\columnwidth]{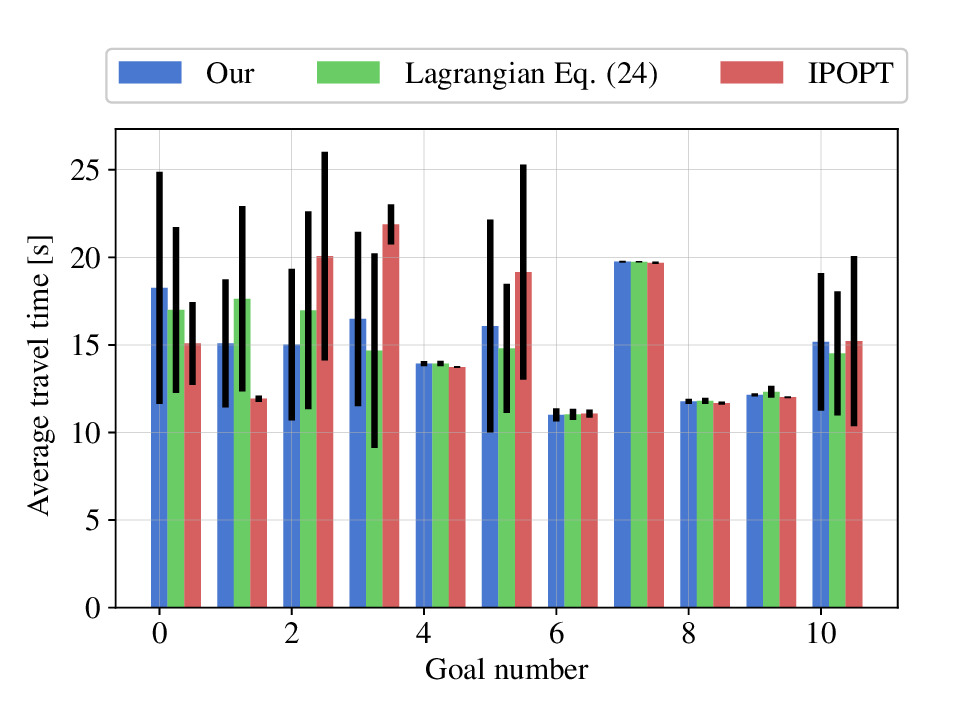}
	\caption{Comparison of the travel time per goal with standard deviation.}
	\label{fig:travel-time}
\end{figure}

To conclude, \figref{fig:travel-time} shows the travel time per goal. It confirms that the three optimization methods are equivalent since the travel time per goal is comparable, as expected.

\section{Conclusion}
\label{sec:conclusion}

In this paper, we presented an extension of factor graphs to constrained optimization based on the \gls{al} method. The method leverages on previous work~\cite{bazzana2023ral} and recent literature on \gls{al} applications~\cite{sodhi2020icra, qadri2022incopt, eriksson2021tpa}. We show its potentials in addressing a variety of constrained optimization problems in a unified framework. The applications shown range from pose estimation to optimal control. We tested our approach in real-world on the pseudo-omnidirectional platform with four steering and driving wheels introduced in~\cite{andreasson2022mdpi}. The experiments suggest that factor graph-solvers can be used to reduce the runtime compared to standard \gls{nlp} methods such as \gls{ipopt}. Open-source code is available at the time of writing. The solver and the factors of the first two applications can be downloaded at \footnotemark. The factors of the \gls{mpc} problem can be found at \footnotemark.

%

\section*{Acknowledgments}
We acknowledge partial financial support from PNRR MUR project PE0000013-FAIR.
\bibliographystyle{plain}
\bibliography{glorified}
\addtocounter{footnote}{-2}
\stepcounter{footnote}\footnotetext{\url{https://gitlab.com/srrg-software/srrg2\_solver/-/tree/smoothed\_constraint\_jacobian?ref_type=heads}}
\stepcounter{footnote}\footnotetext{\url{https://gitlab.com/srrg-software/srrg2\_splam/-/tree/smoothed\_constraint\_jacobian?ref_type=heads}}
\end{document}